\documentclass{article}

\usepackage[final]{neurips_2025}

\usepackage[utf8]{inputenc} 
\usepackage[T1]{fontenc}    
\usepackage{url}            
\usepackage{booktabs}       
\usepackage{amsfonts}       
\usepackage{nicefrac}       
\usepackage{microtype}      
\usepackage{caption}
\usepackage{graphicx, float}
\graphicspath{{images/}}
\usepackage[table]{xcolor}  
\usepackage{bigstrut}    
\usepackage{adjustbox}
\usepackage{amsmath}
\usepackage{hyperref}

\title{LOBERT: Generative AI Foundation Model for Limit Order Book Messages}

\author{%
  Eljas Linna \\
  Data Science Research Centre\\
  Tampere University\\
  \texttt{eljas.linna@tuni.fi} \\
  \And
  Kestutis Baltakys \\
  Data Science Research Centre\\
  Tampere University\\
  \texttt{kestutis.baltakys@tuni.fi} \\
  \And
  Alexandros Iosifidis\\
  Data Science Research Centre\\
  Tampere University\\
  \texttt{alexandros.iosifidis@tuni.fi} \\
  \And
  Juho Kanniainen \\
  Data Science Research Centre\\
  Tampere University\\
  \texttt{juho.kanniainen@tuni.fi} \\
}

\begin{document}

\maketitle

\begin{abstract}
Modeling the dynamics of financial Limit Order Books (LOB) at the message level is challenging due to irregular event timing, rapid regime shifts, and the reactions of high-frequency traders to visible order flow. Previous LOB models require cumbersome data representations and lack adaptability outside their original tasks, leading us to introduce \textsc{LOBERT}, a general-purpose encoder-only foundation model for LOB data suitable for downstream fine-tuning. LOBERT adapts the original BERT architecture for LOB data by using a novel tokenization scheme that treats complete multi-dimensional messages as single tokens while retaining continuous representations of price, volume, and time. With these methods, LOBERT achieves leading performance in tasks such as predicting mid-price movements and next messages, while reducing the required context length compared to previous methods.
\end{abstract}


\section{Introduction}
Financial markets are among the most demanding environments for prediction. They are highly volatile, events unfold on nanosecond time scales, and real capital is at stake. In principle, financial markets should be unpredictable except at the very shortest horizons. This is precisely the regime we study. We develop a novel transformer based model that learns the intricate grammar of limit order book LOB message data. The model directly predicts future LOB messages and enables accurate and fast prediction of downstream LOB features at millisecond resolution, meeting the low latency requirements of high frequency trading.

Accurate and fast modeling of high frequency message flow and LOB features is crucial, as these quantities underpin market making, arbitrage strategies, and market impact modeling for optimal execution. A realistic model of LOB messages can be viewed as capturing the underlying generative process from which observed market dynamics emerge. From this perspective, modeling LOB messages lies at the core of modeling financial markets.

However, message level modeling of LOB markets poses unique challenges compared with other domains commonly addressed by sequence models, such as natural language, speech, or video. While some of the other domains also involve temporal structure, financial markets are characterized by event driven dynamics in which inter event times are extremely irregular and may vary by several orders of magnitude. Moreover, single messages can trigger rapid regime shifts with cascading effects and local regime shifts, which are further amplified by interactions among algorithmic agents that react to each other through the observable message flow. Market making and trading algorithms are designed to respond immediately to the actions of other algorithms as expressed through LOB messages. As key market participants deploy new trading models, the market dynamics themselves evolve, requiring existing models to be frequently updated or even fundamentally redesigned. Parametric mathematical modeling of this process would require detailed knowledge of the internal logic, heterogeneity, and time evolving behavior of these algorithms, which are among the most closely guarded secrets in financial markets.

Machine learning based approaches have demonstrated strong performance in predicting short-horizon mid-price dynamics from LOB data \cite{tran2018temporal, zhang2019deeplob, passalis2021forecasting}. However, most existing methods rely on matrix-style snapshots of the LOB state (Level II data) rather than the underlying message stream (Level III data). 

Recently, a small body of research has proposed modeling order flow as a language, where market messages are tokenized analogously to words in NLP, enabling next-message prediction from historical context. This approach is promising, and it is not surprising that it can generate empirically plausible scenarios of LOB dynamics \cite{nagy2023generative, li2025mars}. In particular, transaction data as well as order submissions and cancellations play a central role in the algorithms of market makers and arbitrage traders. However, existing generative approaches \cite{nagy2023generative, li2025mars} obtain forecasts by autoregressively simulating future order messages and updating the market state through a simulated clearing house. While realistic, this procedure requires running model inference once per predicted event and repeatedly invoking market simulation logic. Consequently, inference latency linearly scales with the prediction horizon, which often needs to be even more more than a hunderd of stems, and is far too slow for real trading environments, where forecasts must be produced within milliseconds. Moreover, these methods split each message into components that must be predicted sequentially (see Appendix \ref{sec:related-work}).

To address these shortcomings, we introduce \textsc{LOBERT}, a general-purpose \emph{encoder-only} foundation model that adapts the highly successful \textsc{BERT} architecture \cite{devlin2019bert} to message-level LOB modeling. In the first stage, the model learns the “grammar” of market microstructure from LOB message data, forming general market representations. In the second stage, the pretrained model is fine-tuned for specific downstream tasks, such as mid-price prediction over arbitrary horizons, using task-specific prediction heads.

The novelty of \textsc{LOBERT} is fourfold. First, we introduce a \emph{one-token-per-message} tokenizer that consolidates side, type, and coarse price and volume cues, while preserving continuous price, volume, and time information via a piecewise linear–geometric scaling. This avoids the token explosion of prior generative models that decompose each message into dozens of sub-tokens. Second, we handle irregular event timing through \emph{continuous-time rotary attention} and a \emph{Masked Message Modeling} objective that jointly masks messages and their surrounding snapshots to prevent label leakage. Third, we propose a \emph{hybrid discrete–continuous} prediction head together with a \emph{Combined} inference scheme, which bounds continuous regressors using token bins and yields higher fidelity than purely discrete or purely continuous outputs. Fourth, we demonstrate strong \emph{efficiency and transferability}: by predicting a single token per message, LOBERT reduces effective sequence length by roughly an order of magnitude on our data and fine-tunes efficiently to diverse downstream tasks using lightweight task-specific heads. To our knowledge, this combination of modeling choices is novel in the context of message-level LOB modeling.


\section{Methodology}
\subsection{Message preprocessing}
\label{subsec:msg-preprocessing}
The main principle behind LOBERT's preprocessing is to produce discrete tokens that represent the most common messages and an approximation of the less common messages, e.g. "new sell order far above the best price with medium volume". This is combined with continuous representation of the information, allowing precisely pinpointing rare values without having to excessively bloat the vocabulary of discrete values.

First, the \textbf{price level} of each message is transformed to stationary price difference measured in ticks from the best opposing price, i.e. the best bid for sell orders and the best ask for buy orders. The price difference is quantized to levels (0, 1, 2, 3, 5, 10) while a duplicate of the price difference is scaled through Piecewise Linear-Geometric Scaling (PLGS, see Appendix \ref{subsec:plg-scaling}) with parameters $\tau_{\text{start}}=10$ ticks, $\tau_{\text{max}}=20$, and $\tau_{\text{clip}}=1000$ ticks. This means the scaling function produces linear output growth up to 10 ticks (50\% of the maximum scaled value), after which growth slows geometrically toward the asymptotic limit. The quantization levels are chosen to distribute the values as evenly as possible across the most common price movements observed in the training data.

Second, the \textbf{volume} is processed with the same method of quantizing to levels (0, 50, 100, 200) and duplicating into a continuous scaled value through PLGS with $\tau_{\text{start}}=200$ units, $\tau_{\text{max}}=400$, and $\tau_{\text{clip}}=1500$ units. The quantization levels are chosen by analyzing the distribution of volumes and observing that volumes are heavily concentrated at round values such as 50, 100, and 200 (over 60\% of all volume values are exactly 100 units in the dataset). We also include an additional binary indicator to denote whether the volume is exactly on the quantized level or between levels ("Y" or "N"), which proves essential for reconstructing the original volume values during inference.

Third, the arrival \textbf{time} of each message is converted to time difference measured from the previous message. Since we did not identify clear repeating patterns in time difference distribution beyond its heavy-tailed nature, we represent it solely with a continuous value scaled through PLGS with parameters $\tau_{\text{start}}=1$ms, $\tau_{\text{max}}=50$, and $\tau_{\text{clip}}=250$ms. The final normalized values sit in a range of [0, 1].

Finally, the \textbf{side} (buy or sell) and \textbf{message type} (new order, edit, delete, execution, hidden order) are combined with the quantized price difference and volume values, along with the volume round indicator, into discrete tokens using a colon delimiter. Together with special tokens for padding, masking, and unknown messages, this process generates 293 distinct message tokens from the training data. A special stringification rule is applied to execution messages (type 4) by converting their price level to 0 since they always occur at the best available price. We now have a vocabulary representing the most important messages, and continuous values supporting them.

\subsection{Order Book snapshot preprocessing}
\label{subsec:lob-preprocessing}
Along with the messages, LOBERT processes a \textbf{snapshot} of the Order Book immediately after the message, i.e., the immediate effect of each message is included in its corresponding snapshot. A single snapshot is an array of 40 values corresponding to the 10 best bid prices and their volumes, and 10 best ask prices and their volumes. Volumes are scaled using an exponential transformation:

\begin{equation}
v_{\text{scaled}} = 1 - e^{-v_{\text{raw}}/k}
\end{equation}

where $v_{\text{raw}}$ is the raw volume at a given price level and $k=2000$ is a scaling parameter that controls the transformation rate. This maps unbounded volume values to $[0,1)$ with diminishing sensitivity to extremely large volumes. Prices are normalized as tick distances from the opposing best quote, clipped at a maximum distance of $D_{\max}=20$ ticks and normalized to $[0,1]$ through division by the maximum distance. This representation captures the relative depth and spread structure of the order book in a compact, normalized form. See Appendix \ref{subsec:lob-preprocessing} for further details.

\subsection{Model architecture, inputs and outputs}

\begin{figure}[h!]
    \centering
    \caption{Simplified architecture diagram highlighting the unique properties of LOBERT}
    \includegraphics[width=0.95\linewidth]{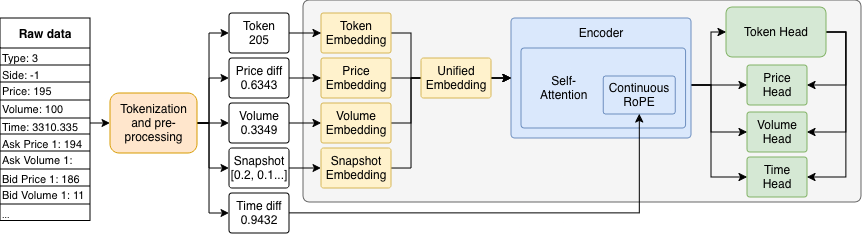}
    \label{fig:architecture}
\end{figure}

LOBERT's architecture (see Figure~\ref{fig:architecture}) builds upon the principles of the original BERT model, implemented using PyTorch. A key innovation is its \emph{multi-modal embedding layer}, which handles the hybrid discrete--continuous nature of LOB data by merging representations from four sources: (1)~discrete message tokens, (2)~continuous price differences, (3)~continuous volume values, and (4)~optionally, LOB snapshot representations integrated through a learned gating mechanism. These embeddings are combined additively to produce a unified representation for each position. In addition to learned positional embeddings, we incorporate continuous Rotary Position Embedding (\textsc{RoPE}) within the self-attention mechanism that operates on cumulative time differences between consecutive messages. This hybrid approach enables the model to capture both sequential ordering and the actual temporal distances between events, addressing the non-uniform nature of market dynamics.

The transformer encoder consists of stacked layers, each containing multi-head self-attention followed by a position-wise feed-forward network with \textsc{Gelu} activation and intermediate dimension expansion. Layer normalization and residual connections are applied after each sub-layer, with dropout ($p=0.1$) for regularization. The model outputs pass through multiple specialized heads: a linear classification head for discrete token prediction (cross-entropy loss), and three regression heads for continuous values (weighted \textsc{Mse} losses). Uniquely, these regression heads employ a \emph{dual-input design}, concatenating both token prediction logits and final hidden states, enabling them to leverage discrete predictions when generating continuous outputs. This architecture facilitates hybrid inference strategies that combine token-based categorical information with refined continuous regression.

\subsection{Training and fine-tuning}

The model’s output structure is also multi-faceted. A standard token prediction head is responsible for predicting the next discrete message token, optimized using a cross-entropy loss function. In parallel, three regression heads are appended after the main transformer layers to predict the continuous values for the given message's price, volume, and time difference. These heads are trained using a Mean Squared Error (MSE) loss, allowing the LOBERT to simultaneously learn the "language" of the most common LOB messages while retaining full granularity.

Our training methodology employs a two-phase strategy designed to first build a robust understanding of the "language" of the LOB message stream and then to refine this understanding for predictive tasks. The initial phase consists of Masked Message Modeling (MMM), analogous to BERT's Masked Language Modeling. In this stage, we train a bidirectional version of LOBERT to reconstruct randomly masked messages within a sequence. To ensure a balanced learning process, the regression losses for price, volume, and time are weighted to be approximately equal after the first epoch of training, while the token prediction loss is given a higher weight to emphasize the importance of identifying the main characteristics of messages. To avoid leaking the masked messages to the model via LOB snapshots around the masked position, the snapshots are also masked for 90\% of positions randomly. Following the MMM phase, we transition to a fine-tuning phase for tasks such as next-message prediction and mid-price prediction. In these phases, all snapshots are visible to the model.


\section{Experiments and results}
In order to demonstrate LOBERT's adaptability, we have conducted fine-tuning experiments for next-message prediction and direct mid-price prediction. Note that due to issues in data access at the time of writing, we could only utilize millisecond-level time differences rather than nanoseconds and thus the time aspect is only partially covered in the experiments. We use 80 trading days of Nasdaq ITCH feed data on AAPL, INTC, MSFT and FB stocks between 2015.05.11 and 2015.09.01 for training, and 10 days between 2015.09.02 and 2015.09.16 for validation. The final 10 days of data from 2015.09.17 to 2015.09.30 are reserved for testing and producing the results discussed here. In total, the dataset has 470M messages divided into 919k non-overlapping sequences of 512 messages each.

For training LOBERT, we use an AdamW optimizer configured with a learning rate of $5 \times 10^{-5}$ and weight decay of $0.01$. Weight decay is disabled for bias terms and layer normalization parameters following standard practice. The learning rate schedule employs Cosine Annealing with Warm Restarts, where the initial restart period $T_0$ is set to $40{,}000$ steps with a multiplicative factor $T_{\text{mult}}=2$ for subsequent cycles, and a minimum learning rate of $5 \times 10^{-6}$. Training is conducted for $10$ epochs with a batch size of $32$, and model evaluation is performed every $15{,}000$ training steps on a held-out validation set. For next-message and mid-price prediction, the model weights are initialized from the pre-trained MMM checkpoint, and the best-performing model is selected based on the lowest validation loss.

\subsection{Next-message prediction experiment}

The pre-trained MMM model is adopted for causal modeling  by using a triangular mask which allows the model to see only past messages, and the label is set to the first message in the hidden section for each sequence. To compare LOBERT with a previous leading model in next-message prediction, we have reconstructed the S5 model as per \cite{nagy2023generative} with 1.2M parameters and without Book Module which processes the LOB snapshots, and trained LOBERT with 1.1M parameters, both with and without Book Module. The models were trained on the same data with roughly an equal amount of training time, and their outputs were transformed with the same processing steps discussed in section \ref{subsec:msg-preprocessing}. The results for individual predicted messages are compared between the two models by measuring the accuracy of their components  (see Table \ref{table:token_accuracy}). The accuracy of LOBERT is considerably higher for each separated component and the full message when considered as a whole.

\begin{table}[htbp]
    \centering
    \caption{Accuracy comparison of predicting the tokenized components of individual messages.}
    \begin{tabular}{l|ccccc}
        \textbf{Model} & \textbf{\shortstack{Message \\ type}} & \textbf{\shortstack{Side or \\ direction}} & \textbf{\shortstack{Price \\ quantized}} & \textbf{\shortstack{Volume \\ quantized}} & \textbf{\shortstack{Full \\ message}} \\
        \midrule
        S5 & 50.8\% & 52.2\% & 18.6\% & 32.1\% & 6.1\% \\
        \textbf{LOBERT} & \textbf{60.6\%} & \textbf{65.1\%} & \textbf{51.8\%} & \textbf{72.1\%} & \textbf{26.4\%} \\
        \textbf{LOBERT with Book Module} & \textbf{61.9\%} & \textbf{65.4\%} & \textbf{53.8\%} & \textbf{72.2\%} & \textbf{27.8\%} \\
        \midrule
    \end{tabular}
    \label{table:token_accuracy}
\end{table}

\begin{figure}[H]
    \centering
    \caption{Value distributions of price, volume, time difference, type and side for LOBERT model with Book Module enabled.}
    \includegraphics[width=0.8\linewidth]{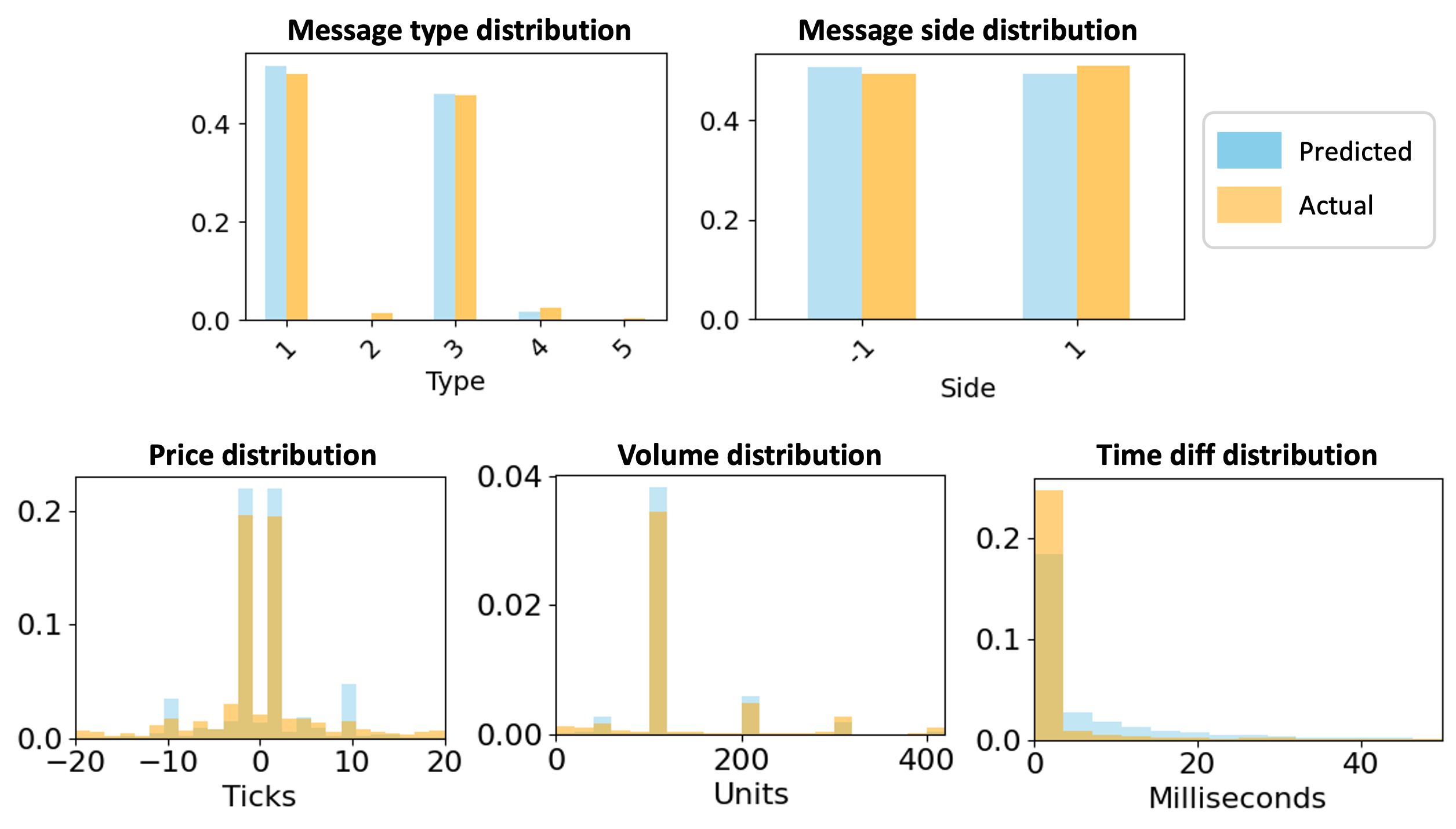}
    \label{fig:distribution}
\end{figure}

To ensure that LOBERT does not overfit to small niches, we inspect the overall distribution of predicted values against the real distribution in the same test dataset (see Figure \ref{fig:distribution}). LOBERT predictions matche the distribution with Pearson correlations of \textit{0.55} for price, \textit{0.37} for volume and \textit{0.52 }for time. One notable deviation stands out at price level 10, which we hypothesize to be caused by conflicts between the token head and price regression head, and we will investigate it further.

\subsection{Message processing comparison}
\label{subsec:processing_comparison}

We compare three message-processing modes that differ in how the discrete token head and the continuous regression heads are utilized at inference time: (i) \textbf{Combined}, which uses the predicted token's quantization range as lower and upper bounds for the regressor; (ii) \textbf{Token only}, which maps the predicted token to its quantized price (bin-start) and volume (bin-centre) while discarding the regressors; and (iii) \textbf{Regressor only}, which ignores the token prediction and uses the three regression outputs directly. Note that even though Time Difference is only predicted with regressor, we include it in the table for reference on performance. To quantify distributional fidelity, we evaluate marginal discrepancies between predictions and ground truth on the test split using the Wasserstein-1 distance (W1), Jensen--Shannon divergence (JSD), and Total Variation distance (TVD) for \emph{Price}, \emph{Volume}, and \emph{Time Difference}. Lower is better for all metrics.

Across all three variables and all three metrics, the \textbf{Combined} mode consistently achieves the smallest discrepancy relative to both ablations (see Table~\ref{table:msg_processing_comparison}). This pattern indicates a complementary effect: the token head captures high-probability structural motifs (e.g., side/type and coarse price--volume quantiles), while the regressors refine continuous residuals and rare values, reducing discretization artefacts present in the token-only pathway and stabilizing the regressor-only pathway. These gains substantiate the design choice of retaining both channels in LOBERT’s output layer.

\begin{table}[h!]
    \centering
    \caption{Distributional discrepancy (lower is better) between predicted and true marginals for \emph{Price}, \emph{Volume}, and \emph{Time Difference}. Reported metrics are Wasserstein-1 (W1), Jensen--Shannon divergence (JSD), and Total Variation distance (TVD). Time difference is only predicted with regressors but we include the results here for reference on performance.}
    \label{table:msg_processing_comparison}
    \setlength{\tabcolsep}{5pt}
    \begin{adjustbox}{max width=\textwidth}
    \begin{tabular}{l|ccc|ccc|ccc}
        & \multicolumn{3}{c|}{\textbf{Price}} & \multicolumn{3}{c|}{\textbf{Volume}} & \multicolumn{3}{c}{\textbf{Time Difference}} \\
        \textbf{Method} & \textbf{W1} & \textbf{JSD} & \textbf{TVD} & \textbf{W1} & \textbf{JSD} & \textbf{TVD} & \textbf{W1} & \textbf{JSD} & \textbf{TVD} \\
        \midrule
        \textbf{Combined} & \textit{\textbf{10.04}} & \textit{\textbf{0.2111}} & \textit{\textbf{0.1839}} & \textit{\textbf{76.86}} & \textit{\textbf{0.1696}} & \textit{\textbf{0.1200}} & \textit{N/A} & \textit{N/A} & \textit{N/A} \\
        Token only                              & \textit{15.44} & \textit{0.5566} & \textit{0.6134} & \textit{85.36} & \textit{0.2446} & \textit{0.1585} & \textit{N/A} & \textit{N/A} & \textit{N/A} \\
        Regressor only                          & \textit{10.85} & \textit{0.3959} & \textit{0.4767} & \textit{86.09} & \textit{0.6623} & \textit{0.8395} & \textit{10.658} & \textit{0.2681} & \textit{0.3270} \\
        \bottomrule
    \end{tabular}
    \end{adjustbox}
\end{table}

\subsection{Mid-price prediction experiment}
We adapt LOBERT for mid-price direction by adding a classification head on top of the MMM-pretrained encoder. The task predicts the direction of the average tick-normalized mid-price movement over a message-based horizon. Let \(m(t)\) be the mid-price at message \(t\), and define the horizon set \(H=\{10,50,100\}\) messages (message-based rather than time-based). We use the average future tick-normalized mid-price
\(\bar{m}_h(t)=\frac{1}{h}\sum_{i=1}^{h} m(t+i)\)
and assign a ternary label
\[
\ell_h(t)=
\begin{cases}
+1, & \bar{m}_h(t)-m(t) > \tau(h),\\
-1, & \bar{m}_h(t)-m(t) < -\tau(h),\\
0, & \text{otherwise},
\end{cases}
\]
where \(\tau(h)\) is a horizon-dependent threshold (in the same units as \(m\)). We choose a logarithmically increasing threshold to reflect the sublinear growth of typical price moves with horizon length:
\[
\tau(h) \;=\; \frac{1}{1000}\,\operatorname{round}\!\Big(100\,\log_{2}\!\tfrac{h}{10}\Big).
\]
The averaging in \(\bar{m}_h(t)\) reduces microstructure noise relative to end-of-horizon labeling.

We train a 3-way cross-entropy head where the inputs include snapshots and messages up to time \(t\), and the target is \(\ell_h(t)\). We report \textit{macro-F1} and \textit{coverage} under \textit{selective prediction}: a prediction is emitted only if the \emph{maximum softmax probability} exceeds a confidence threshold \(\tau_c\in\{0.3,0.4,\ldots,0.9\}\). Results are averaged over four securities (AAPL, INTC, MSFT, FB). As a baseline, we re-train \textsc{DeepLOB}~\cite{zhang2019deeplob} under the same data and horizons.

Table~\ref{table:midprice} shows that LOBERT consistently outperforms or matches DeepLOB across horizons and confidence thresholds. We also observe that LOBERT is more responsive to confidence filtering: as \(\tau_c\) increases from 0.3 to 0.9, macro-F1 improves markedly (e.g., \(H=100\): \(0.55\rightarrow 0.88\)) while coverage decreases (1.00 \(\rightarrow\) 0.10), indicating effective ranking of easy vs. hard cases. See Appendix \ref{sec:midprice-graphs} for asset-specific curves.

\begin{table}[htbp]
  \centering
  \caption{Mid-price prediction F1 metrics (selective F1 at a confidence threshold) and coverage for prediction horizons of 10, 50, and 100. The F1 and coverage values are averaged over four securities: FB, INTC, MSFT, and AAPL. Detailed results are provided in the Appendix. Note that selective F1 should be read as the model’s F1 score only on the cases where it was confident enough to make a prediction.}
  \label{table:midprice}

  \resizebox{\textwidth}{!}{
  \begin{tabular}{r|rr|rr|rr|rr|rr|rr}
        & \multicolumn{4}{c|}{$H = 10$}   & \multicolumn{4}{c|}{$H = 50$}   & \multicolumn{4}{c}{$H = 100$} \\
        Confidence & \multicolumn{2}{c}{LOBERT} & \multicolumn{2}{c|}{DeepLOB} & \multicolumn{2}{c}{LOBERT} & \multicolumn{2}{c|}{DeepLOB} & \multicolumn{2}{c}{LOBERT} & \multicolumn{2}{c}{DeepLOB} \\
    \multicolumn{1}{l|}{threshold} & \multicolumn{1}{l}{F1} & \multicolumn{1}{l}{Cover.} & \multicolumn{1}{l}{F1} & \multicolumn{1}{l|}{Cover.} & \multicolumn{1}{l}{F1} & \multicolumn{1}{l}{Cover.} & \multicolumn{1}{l}{F1} & \multicolumn{1}{l|}{Cover.} & \multicolumn{1}{l}{F1} & \multicolumn{1}{l}{Cover.} & \multicolumn{1}{l}{F1} & \multicolumn{1}{l}{Cover.} \bigstrut[b]\\
    \hline
    0.3   & \cellcolor[rgb]{.98,.737,.749}\textit{0.51} & \cellcolor[rgb]{.353,.541,.776}1.00 & \cellcolor[rgb]{.973,.412,.42}0.44 & \cellcolor[rgb]{.353,.541,.776}1.00 & \cellcolor[rgb]{.984,.984,.996}\textit{0.56} & \cellcolor[rgb]{.353,.541,.776}1.00 & \cellcolor[rgb]{.98,.839,.851}0.53 & \cellcolor[rgb]{.353,.541,.776}1.00 & \cellcolor[rgb]{.984,.949,.961}\textit{0.55} & \cellcolor[rgb]{.353,.541,.776}1.00 & \cellcolor[rgb]{.984,.941,.953}0.55 & \cellcolor[rgb]{.353,.541,.776}1.00 \bigstrut[t]\\
    0.4   & \cellcolor[rgb]{.98,.78,.792}\textit{0.52} & \cellcolor[rgb]{.443,.604,.808}0.95 & \cellcolor[rgb]{.973,.42,.431}0.45 & \cellcolor[rgb]{.388,.565,.788}0.98 & \cellcolor[rgb]{.976,.984,.988}\textit{0.56} & \cellcolor[rgb]{.427,.592,.804}0.96 & \cellcolor[rgb]{.984,.851,.863}0.53 & \cellcolor[rgb]{.38,.561,.788}0.99 & \cellcolor[rgb]{.984,.988,.996}\textit{0.56} & \cellcolor[rgb]{.431,.596,.804}0.96 & \cellcolor[rgb]{.984,.949,.961}0.55 & \cellcolor[rgb]{.365,.549,.78}0.99 \\
    0.5   & \cellcolor[rgb]{.984,.949,.957}\textit{0.55} & \cellcolor[rgb]{.753,.824,.918}0.79 & \cellcolor[rgb]{.973,.475,.482}0.46 & \cellcolor[rgb]{.541,.675,.843}0.90 & \cellcolor[rgb]{.91,.957,.933}\textit{0.60} & \cellcolor[rgb]{.784,.847,.929}0.77 & \cellcolor[rgb]{.984,.878,.89}0.54 & \cellcolor[rgb]{.553,.682,.847}0.89 & \cellcolor[rgb]{.902,.953,.925}\textit{0.60} & \cellcolor[rgb]{.792,.851,.933}0.77 & \cellcolor[rgb]{.988,.988,1}0.56 & \cellcolor[rgb]{.514,.655,.835}0.92 \\
    0.6   & \cellcolor[rgb]{.902,.957,.925}\textit{0.60} & \cellcolor[rgb]{.984,.929,.941}0.61 & \cellcolor[rgb]{.976,.565,.573}0.47 & \cellcolor[rgb]{.792,.851,.933}0.77 & \cellcolor[rgb]{.773,.902,.816}\textit{0.67} & \cellcolor[rgb]{.98,.827,.839}0.51 & \cellcolor[rgb]{.98,.776,.784}0.52 & \cellcolor[rgb]{.918,.941,.976}0.70 & \cellcolor[rgb]{.757,.898,.804}\textit{0.68} & \cellcolor[rgb]{.98,.792,.804}0.48 & \cellcolor[rgb]{.984,.957,.969}0.55 & \cellcolor[rgb]{.984,.984,.996}0.66 \\
    0.7   & \cellcolor[rgb]{.769,.902,.812}\textit{0.68} & \cellcolor[rgb]{.98,.792,.8}0.47 & \cellcolor[rgb]{.976,.643,.655}0.49 & \cellcolor[rgb]{.988,.988,1}0.67 & \cellcolor[rgb]{.604,.835,.671}\textit{0.77} & \cellcolor[rgb]{.976,.647,.659}0.33 & \cellcolor[rgb]{.984,.867,.878}0.53 & \cellcolor[rgb]{.984,.918,.929}0.60 & \cellcolor[rgb]{.596,.831,.663}\textit{0.77} & \cellcolor[rgb]{.976,.604,.616}0.29 & \cellcolor[rgb]{.984,.957,.969}0.55 & \cellcolor[rgb]{.984,.851,.859}0.53 \\
    0.8   & \cellcolor[rgb]{.627,.843,.69}\textit{0.75} & \cellcolor[rgb]{.976,.69,.698}0.37 & \cellcolor[rgb]{.976,.667,.675}0.49 & \cellcolor[rgb]{.984,.906,.914}0.58 & \cellcolor[rgb]{.443,.769,.529}\textit{0.86} & \cellcolor[rgb]{.973,.541,.549}0.23 & \cellcolor[rgb]{.984,.988,.996}0.56 & \cellcolor[rgb]{.98,.824,.831}0.50 & \cellcolor[rgb]{.459,.776,.545}\textit{0.85} & \cellcolor[rgb]{.973,.498,.506}0.18 & \cellcolor[rgb]{.953,.976,.969}0.58 & \cellcolor[rgb]{.98,.722,.733}0.41 \\
    0.9   & \cellcolor[rgb]{.502,.792,.58}\textit{0.82} & \cellcolor[rgb]{.976,.604,.612}0.29 & \cellcolor[rgb]{.976,.604,.612}0.48 & \cellcolor[rgb]{.98,.816,.824}0.50 & \cellcolor[rgb]{.478,.784,.561}\textit{0.84} & \cellcolor[rgb]{.973,.455,.463}0.14 & \cellcolor[rgb]{.918,.961,.937}0.60 & \cellcolor[rgb]{.98,.722,.729}0.40 & \cellcolor[rgb]{.388,.745,.482}\textit{0.88} & \cellcolor[rgb]{.973,.412,.42}0.10 & \cellcolor[rgb]{.898,.953,.922}0.61 & \cellcolor[rgb]{.976,.6,.608}0.28 \\
  \end{tabular}
  } 
\end{table}

LOBERT's inference is currently slower than DeepLOB's by 53\%. However, Transformer-based architectures such as LOBERT have a large arsenal of potential mechanisms to increase throughput by several multiples, which we plan to implement in future versions. See Appendix \ref{sec:inference-speed} for details on inference speed.


\section{Discussion and conclusion}

Our work on LOBERT demonstrates the significant potential of adapting methodologies from the world of natural language processing to model the dynamics of the LOB. By treating the stream of LOB events as a unique language, we leverage the well-established pattern-recognition capabilities of the BERT architecture. The initial results from our next-message and mid-price prediction experiments are promising, suggesting that a model can indeed learn the intricate "grammar" of market data. Thus, LOBERT foundation model can be adapted to countless use cases with a small amount of fine-tuning, opening the door for a completely new way of creating models in LOB domain, similar to how the original BERT model changed language modeling.

Despite the advancements, we acknowledge the limitations in our current work. Namely, our experiments so far do not guarantee the model's ability to produce realistic long-term sequences. Our future work will validate this capability through extensive sequence-level analysis. Furthermore, we have not yet completed the study on the breadth of task types which LOBERT can be adapted to. LLMs are also known to struggle with tracking the state of entities through changes, and LOBERT may face a similar challenge in tracking the size and location of orders on the book when the price levels move. Our current implementation simplifies this by not modeling individual order IDs. Additionally, the computational cost and inference time of the transformer architecture are non-trivial.

In conclusion, LOBERT represents a promising new direction for modeling LOB dynamics. By adapting the BERT framework, we have developed a model capable of understanding the fundamental language of market messages, unifying discrete and continuous event data into a single, cohesive framework. We will now focus on addressing the current limitations and expanding the model's capabilities. Before the conference, our roadmap includes conducting comprehensive sequence prediction experiments to validate the model's generative quality and expanding the fine-tuning coverage study. The performance of the model will be improved with more efficient attention mechanisms and enhanced training techniques introduced in later adaptations of BERT. To further enrich the model's understanding, we will incorporate crucial market state information, such as the time of day, the current bid-ask spread, and full LOB snapshots.

\section*{Acknowledgments}
The authors thank the Nasdaq Nordic Foundation for financial support through the project "Deep Generative Models for Limit Order Book Markets with Applications to Optimal Execution and Spoofing Detection”.

\bibliographystyle{plain}
\bibliography{references}

\appendix

\section{Related work}
\label{sec:related-work}
Previous attempts to model the LOB have spanned a range of machine learning techniques. Early efforts focused on supervised learning, framing the problem as a classification task to predict mid-price movements \cite{sirignano2021universal,tran2018temporal,tsantekidis2017forecasting,zhang2019deeplob}. More recently, GANs have been applied for LOB snapshot prediction \cite{cont2023limit}, and generative models such as \cite{nagy2023generative} based on sequence modeling as with \cite{smith2022simplified} by splitting each message into 22 tokens and predicting them one by one. \cite{wheeler2024marketgpt} presents a generative pre-trained transformer for financial data simulation but it is also slowed down by splitting messages into 24 tokens. \cite{berti2025trades} introduces a transformer-based diffusion engine for LOB simulations, offering flexible generation but with the cost of heavy computational requirement by sampling requires hundreds of iterative steps. \cite{berti2025tlob} proposes a dual-attention transformer for price trend prediction, improving accuracy, yet focusing narrowly on trend forecasting rather than full LOB message generation. In a concurrent line of work, \cite{li2025mars} introduces MarS, a financial market simulation engine powered by a generative foundation model (LMM) that also operates at the order level to create interactive and controllable market simulations.

\section{Piecewise Linear-Geometric Scaling}
\label{subsec:plg-scaling}
To handle the wide dynamic range of LOB features while preserving both precision at small values and representation of extreme values, we employ Piecewise Linear-Geometric Scaling. For a raw input value $x$ (measured in original units: ticks for price, units for volume, or milliseconds for time), with parameters $\tau_{\text{start}}$ (transition point in input units), $\tau_{\text{max}}$ (maximum scaled output value), and $\tau_{\text{clip}}$ (input clipping threshold), the transformation proceeds in two stages.

First, we compute an intermediate scaled value $s(x)$:

\begin{equation}
s(x) = \begin{cases}
x & \text{if } x \leq \tau_{\text{start}} \\
\tau_{\text{start}} + \sum_{k=0}^{\lfloor x - \tau_{\text{start}} \rfloor} \mu^k & \text{if } \tau_{\text{start}} < x \leq \tau_{\text{clip}} \\
\tau_{\text{start}} + \sum_{k=0}^{\lfloor \tau_{\text{clip}} - \tau_{\text{start}} \rfloor} \mu^k & \text{if } x > \tau_{\text{clip}}
\end{cases}
\end{equation}

where the geometric decay factor $\mu$ is determined by requiring that the infinite series converges to $\tau_{\text{max}}$:

\begin{equation}
\mu = 1 - \frac{1}{\tau_{\text{max}} - \tau_{\text{start}}}
\end{equation}

This ensures that $\lim_{x \to \infty} s(x) = \tau_{\text{start}} + \frac{1}{1-\mu} = \tau_{\text{max}}$.

Second, we normalize the intermediate scaled value to the range $[0,1]$:

\begin{equation}
x_{\text{scaled}} = \frac{s(x)}{\tau_{\text{max}}}
\end{equation}

This two-stage process provides several advantages: (i) linear growth for $x \leq \tau_{\text{start}}$ preserves sensitivity to small, frequent values; (ii) geometric decay for $x > \tau_{\text{start}}$ compresses extreme outliers; (iii) the ratio $\tau_{\text{start}}/\tau_{\text{max}}$ directly controls the proportion of the output range [0,1] dedicated to linear scaling; and (iv) asymptotic convergence prevents unbounded values from dominating the learned representations. For continuous input values between discrete steps, linear interpolation is applied within each unit interval.

\section{Snapshot scaling}

The volume at each price level is transformed using an exponential scaling function to normalize the distribution:

\begin{equation}
v_{\text{scaled}} = 1 - e^{-v_{\text{raw}}/k}
\end{equation}

where \(v_{\text{raw}}\) is the raw volume at a given price level and \(k\) is a scaling parameter that controls the transformation rate. This transformation maps unbounded volume values to the range \([0, 1)\), with the exponential function providing diminishing sensitivity to extremely large volumes.

Prices are normalized by representing price levels as distances from the best opposite quote. The distance from each price level to the best opposite quote is normalized to capture relative position in the order book. For bid prices:

\begin{equation}
d_{\text{bid},i} = \frac{\min\left(\left\lfloor\frac{p_{\text{ask}}^{(1)} - p_{\text{bid}}^{(i)}}{\Delta p}\right\rceil - 1, D_{\max}\right)}{D_{\max}}
\end{equation}

where \(p_{\text{ask}}^{(1)}\) is the best ask price, \(p_{\text{bid}}^{(i)}\) is the \(i\)-th bid price level, \(\Delta p\) is the minimum tick size, and \(D_{\max}\) is the maximum distance threshold (clip\_dist). The term \(\left\lfloor \cdot \right\rceil\) denotes rounding to the nearest integer.

Similarly, for ask prices:

\begin{equation}
d_{\text{ask},i} = \frac{\min\left(\left\lfloor\frac{p_{\text{ask}}^{(i)} - p_{\text{bid}}^{(1)}}{\Delta p}\right\rceil - 1, D_{\max}\right)}{D_{\max}}
\end{equation}

where \(p_{\text{bid}}^{(1)}\) is the best bid price and \(p_{\text{ask}}^{(i)}\) is the \(i\)-th ask price level. Both distance metrics are clipped to \([0, 1]\) through the minimum operation and subsequent normalization by \(D_{\max}\).

\section{Inference speed}
\label{sec:inference-speed}
The present implementation of LOBERT prioritizes predictive quality over latency. On a single NVIDIA V100 GPU, the model attains a throughput of \textbf{281.87} predictions per second, corresponding to \textbf{47\%} of \textsc{DeepLOB}'s throughput under the same evaluation conditions with batch size of 1. While these figures reflect a quality-focused design choice, the Transformer backbone offers several well-understood avenues for accelerating inference without altering task formulation.

\textbf{Quantization.} Reducing arithmetic precision from FP32 to lower-bit formats (e.g., INT8) can substantially increase throughput by improving compute and memory efficiency. In prior work on Transformer encoders, quantization has yielded \emph{up to $4\times$} speed-ups with minimal accuracy impact on supported hardware \cite{kim2021bert}. Applying post-training or quantization-aware variants to LOBERT is therefore a promising first lever for latency reduction.

\textbf{Structured pruning.} Structured removal of computation---such as pruning full layers, attention heads, or blocks within the feed-forward sublayers---reduces end-to-end FLOPs and wall-clock time while preserving dense tensor layouts that deploy efficiently. Reported gains reach \emph{up to $2\times$} at modest accuracy cost when pruning is paired with brief recovery fine-tuning \cite{xia2022}. Given LOBERT's compact depth, conservative layer- or width-pruning regimes are natural candidates. In this line, structured pruning adapters can be utilized for creating specialized models with low parameter count \cite{hedegaard2024structured}.

\textbf{Early-exit strategies.} Adaptive inference equips intermediate layers with auxiliary classifiers and halts computation once confidence is sufficient. On average, such mechanisms have achieved \emph{$\sim1.57\times$} speed-ups, with the actual gain governed by the distribution of “easy” inputs and the exit threshold \cite{zhou2020,bakhtiarnia2022single}. Integrating early exits into LOBERT preserves the model’s capacity for difficult cases while trimming work on routine ones.

\textbf{Efficient attention.} For longer input windows, the quadratic cost of standard self-attention becomes a bottleneck. Efficient attention formulations that reduce the effective complexity can deliver \emph{up to $1.5\times$} speed-ups at moderate sequence lengths while maintaining predictive quality \cite{wang2020}. These methods are especially relevant if future deployments require extending the historical context processed per inference step.

\textbf{Efficient online operation.} All above-described approaches are commonly used to lower the computational cost in \textit{static inference} settings, i.e., when a sequence segment is introduced to the model for obtaining the corresponding inference result. Even though adopting them would lead to a higher throughput, the fact that LOBERT needs to be combined with a temporal sliding window for operating on LOB data streams leads to high computational redundancies. This is a traditional limitation of Transformer encoder models compared to models formed by layers having a recurrent formulation, as are the LSTM layers of \textsc{DeepLOB}. However, recent reformulation of Transformers as Continual Inference Networks \cite{Hedegaard2022Continual} has led to the so-called Continual Transformers \cite{Hedegaard2023Continual} and more recently to the Deep Continual Transformers \cite{hedegaard2025DeepCot} opening up the possibility of obtaining redundancy-free continual LOBERT reaching speed ups of up to two orders of magnitude, and with negligible computational cost increase for increased input window sizes.

\section{Continuous Rotary Position Embedding}
We extend the standard Rotary Position Embedding (RoPE) mechanism introduced in \cite{su2024} to handle continuous time by replacing discrete position indices with cumulative time differences. For each attention head with dimension $d_h$, we define frequency components $\theta_i = 10000^{-2i/d_h}$ for $i \in \{0, 1, \ldots, d_h/2-1\}$. Given cumulative time $t \in \mathbb{R}^+$ at each sequence position, we compute rotation angles $\phi_i(t) = t \cdot \theta_i$. The continuous RoPE transformation is then applied to query and key vectors $\mathbf{q}, \mathbf{k} \in \mathbb{R}^{d_h}$ by rotating consecutive dimension pairs: for each pair $(q_{2i}, q_{2i+1})$, we compute $(q'_{2i}, q'_{2i+1}) = (q_{2i}\cos\phi_i(t) - q_{2i+1}\sin\phi_i(t), q_{2i}\sin\phi_i(t) + q_{2i+1}\cos\phi_i(t))$, and analogously for $\mathbf{k}$. This formulation preserves the relative position encoding property of RoPE while naturally accommodating irregular time intervals between events in the Limit Order Book message stream.

\section{Asset specific F1 and coverage scores with confidence thresholding}
\label{sec:midprice-graphs}
\begin{figure}[H]
    \centering
    \includegraphics[width=0.95\linewidth]{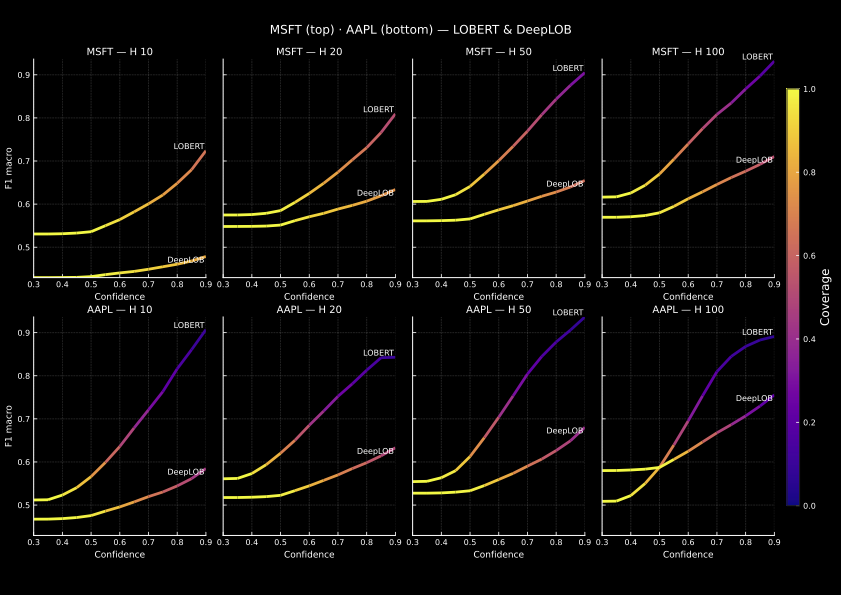}
    \caption{Comparison of LOBERT and DeepLOB for mid-price prediction plotted by confidence threshold, F1 macro score and coverage for MSFT and AAPL assets.}
    \label{fig:asset-specific-scores-msft-aapl}
\end{figure}

\begin{figure}
    \centering
    \includegraphics[width=0.95\linewidth]{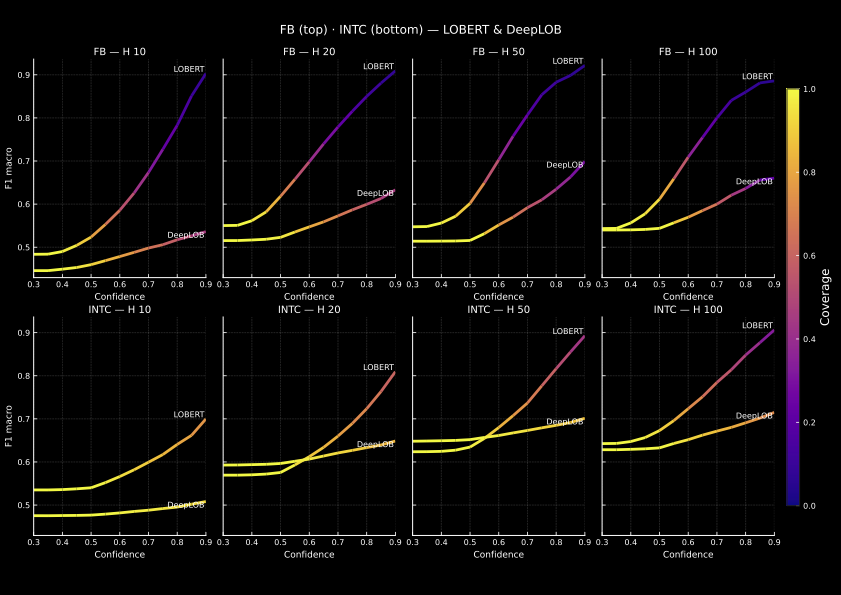}
    \caption{Comparison of LOBERT and DeepLOB for mid-price prediction plotted by confidence threshold, F1 macro score and coverage for FB and INTC assets.}
    \label{fig:asset-specific-scores-fb-intc}
\end{figure}

Figures \ref{fig:asset-specific-scores-msft-aapl} and \ref{fig:asset-specific-scores-fb-intc} describe how the F1 score of predictions increases as we increase the minimum threshold required for action. LOBERT reacts more aggressively to confidence thresholds, as seen in the steep upwards curvature, while DeepLOB's confidence is more uniform. At the same time, the coverage of LOBERT decreases.

\end{document}